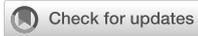

# Yes we care!-Certification for machine learning methods through the care label framework

Katharina J. Morik[1]\*, Helena Kotthaus[1], Raphael Fischer[1], Sascha Mücke[1], Matthias Jakobs[1], Nico Piatkowski[2], Andreas Pauly[1], Lukas Heppe[1] and Danny Heinrich[1]

[1]Faculty of Computer Science, TU Dortmund University, Dortmund, Germany, [2]Fraunhofer Institute for Intelligent Analysis and Information Systems, Sankt Augustin, Germany

Machine learning applications have become ubiquitous. Their applications range from embedded control in production machines over process optimization in diverse areas (e.g., traffic, finance, sciences) to direct user interactions like advertising and recommendations. This has led to an increased effort of making machine learning trustworthy. Explainable and fair AI have already matured. They address the knowledgeable user and the application engineer. However, there are users that want to deploy a learned model in a similar way as their washing machine. These stakeholders do not want to spend time in understanding the model, but want to rely on guaranteed properties. What are the relevant properties? How can they be expressed to the stakeholder without presupposing machine learning knowledge? How can they be guaranteed for a certain implementation of a machine learning model? These questions move far beyond the current state of the art and we want to address them here. We propose a unified framework that certifies learning methods *via* care labels. They are easy to understand and draw inspiration from well-known certificates like textile labels or property cards of electronic devices. Our framework considers both, the machine learning theory and a given implementation. We test the implementation's compliance with theoretical properties and bounds.

## 1. Introduction

Machine learning (ML) has become the driving force pushing diverse computational services like search engines, robotics, traffic forecasting, natural language processing, and medical diagnosis, to only mention a few. This has led to a more diverse group of people affected by ML.

Placing the human in the center of ML investigates the needs of the user and the interaction between developer and user. Many approaches refer—possibly indirectly—to a pairing of a developer and a deployer, interacting for a certain application. Typical examples are from ML in sciences, where the physicist, biologist, or drug developer interacts with the ML expert to establish a reliable data analysis process. In the long run, others may benefit from this without ever accessing the ML process. Patients, for instance, take for granted that diagnosis methods or drugs are approved by a valid procedure.

Other applications involve further parties. A vendor company, its online shop, the company that optimizes the click rate, a recommendation engine, and the customers buying the products are all playing their part in modern sales ecosystems. Modern financial business processes, like money laundry detection, involve a network of stakeholders who need to know the reliability of the ML classifications. Companies apply diverse ML methods, and some have employees who know ML well. For them, to inspect a learned model is important and explainable AI is serving them. In contrast, the customers are affected by the process but do not face the ML system directly nor do they interact with its developers. They rely on tech companies to have done their job in a trustworthy manner.

In an even broader context, societies establish regulations that protect individual rights, e.g., regarding privacy of data and fair business processes. Moreover, the goals of sustainability and the fight against climate change demand regulations on energy consumption of ML processes. Now the regulating agencies need valid information about ML processes.

We see the diversity of stakeholders who need valid information in order to accept or not accept a certain ML process. Some of them know the ML theory and are experienced in evaluating models. Some of them will interpret the models directly if they are nicely visualized. Some of them will find the time for an interactive inspection of models. All these needs have raised considerable attention in the AI community and paved the way for explainable AI (XAI). However, methods that inform users who do not want to spend time in learning about ML method are missing — this is the type of user we want to address.

We consider this user type a customer of ML who wants the product to fulfill some requirements. Whether a method meets the expectation is partially given by theoretical properties. However, such statements are scattered across decades of scientific publications, and finding and understanding them requires years of studies. Of course, our 'customers' who do not want to invest time into considering a particular model will have even less interest in visiting courses. The necessity of communicating theoretical insights more easily has led to the concept of care labels (Morik et al., 2021), which we adopt here. This concept moves beyond individual man-machine interaction toward a public declaration of a method's properties. It is closer to certification than to XAI and would allow for specific regulations, for instance, regarding energy consumption.

The design of ML care labels requires defining the set of relevant properties. Robustness in the sense that small changes to data should not deteriorate the model too much is a property that is studied intensively. Runtime and memory bounds are straightforward. Communication needs are important for applications in the area of the Internet of Things. Energy consumption is important due to the potentially high impact on our environment (Strubell et al., 2020). As an example, the state-of-the-art NLP model "BERT" has an average power consumption of 12 kW, with training alone consuming 1 MWh (as much as a single-person household consumes in 8 months[1]).

Most classes of ML methods, like exponential families, offer a range of algorithms for training and inference. Consider the choice of algorithm for performing inference on probabilistic graphical models, which leads to totally different theoretical properties, runtimes, and $CO_2$ footprints. The marginal probabilities are often under- or overestimated by using the approximative loopy belief propagation (LBP) algorithm instead of the exact junction tree (JT) algorithm, which on the downside has high asymptotic runtime complexity. If the implications of choosing among both algorithms is indicated clearly by a care label, even an inexperienced user can decide whether the particular application requires exact JT or resource-friendly LBP inference. In general, different ML methods may need different categories, and even the same category may need different criteria to be tested. Hence, if we take a more detailed look at the overall ML field, there is not one single set of categories with test criteria for all. Instead, an expert database should store the specific instances of the categories for ML methods.

Moreover, considering static characteristics of a method does not suffice. Worst-case asymptotic time and memory bounds are given by theory, but can vary by orders of magnitude across compute platforms, even if they implement the same abstract method. A convolutional neural network (CNN), for example, may be trained on a resource-hungry GPU system consuming several hundred Watt, or on a microcontroller which usually consume less than W (e.g., Arduino or field programmable gate array (FPGA)). The latter, on the other hand, may be severely constrained with respect to the number of layers, or input data types: An FPGA may work best using only integer arithmetic, or the Arduino may only have 256 kB of RAM, which limits the model's number of parameters. In general, the same method can be implemented on different hardware architectures and particular implementations might vary. Hence, the more dynamic behavior of ML execution environments must be covered by the labels, as well.

For each property, ranges of values need to be defined that will then be expressed by symbols similar to those on the paper slips found in clothes and textiles. Since we want to validate the properties, we need criteria which classify a certain instance of a method into the appropriate value range. Where static properties may be listed based on theoretical results, the dynamic properties of a particular implementation on particular hardware demand tests on specific data sets. Overall, for the set of properties with their value ranges, a certification process needs to be implemented.

---

1 Based on the average power consumption per capita in Europe https://ec.europa.eu/eurostat/statistics-explained/index.php/Electricity_and_heat_statistics.

In this work, we propose a novel means of communication between ML scientists and stakeholders that goes beyond logical, visual or natural language descriptions of single models. We instead aim at providing a framework for certifying ML methods in general, as schematically displayed in Figure 1. Our contribution comprises the following points:

1. We present an easy-to-understand *care label design*, serving as a single graphical certificate (Figure 2) for ML methods and their implementations.
2. We devise a *rating* system, drawing from an *expert knowledge database* created, maintained and continually expanded by the research community.
3. We introduce *categories* under which we bundle *criteria*, which represent important properties of ML methods. They are stored in the expert knowledge base.
4. We suggest to certify a given implementation against its underlying theory with the help of *reliability and performance bound checks* on *profiling data sets*, and reporting resource consumption.
5. We define *badges* that are awarded to ML methods that fulfill certain noteworthy criteria.
6. We present a concept for a *Certification Suite* that accesses the expert knowledge database and certifies a method together with its implementation.

We start by giving an overview of good standards that have already been achieved for ML certification, and identifying their shortcomings. Next up, we introduce our novel care label concept and its constituting parts in Section 3. In Section 4 we put our concept into practice for Markov random fields (MRFs), a class of probabilistic graphical models with a wide range of applications. MRFs constitute a powerful probabilistic tool with a rich theoretical background, serving to illuminate all aspects of our care label concept. We conclude our work with a summary of our investigations, and outline future work in Section 5.

## 2. Related work

The importance of trustworthy AI methods is increasing, especially because decision-making takes data-based models more and more into account (Bellotti and Edwards, 2001; Floridi et al., 2018; Lepri et al., 2018; Houben et al., 2021). In her comprehensive book, Virginia Dignum addresses AI's ethical implications of interest to researchers, technologists, and policymakers (Dignum, 2019). Another recent book brings together many perspectives of AI for humanity and justifies the urgency of reflecting on AI with respect to reliability, accountability, societal impact, and juridical regulations (Braunschweig and Ghallab, 2021). Brundage et al. (2020) and Langer et al. (2021) summarized important aspects of developing trustworthy learning systems. Their reports emphasize that institutional mechanisms (e.g., auditing, red team exercises, etc.), software mechanisms (e.g., audit trails, interpretability, etc.) and hardware mechanisms (assessing secure hardware, high precision compute management, etc.) are required for obtaining trusted systems, and that the diversity of stakeholders needs to be taken into account. This brings the issue of certification and testing to the foreground (Cremers et al., 2019). Miles Brundage and colleagues argue that descriptions must be verifiable claims (Brundage et al., 2020). Verification methods have been applied for trustworthy deep neural networks (Huang et al., 2020) and for investigate verification of probabilistic real-time systems (Kwiatkowska et al., 2011).

Where privacy-preserving data mining (e.g., Atzori et al., 2008) on the side of data analysis methods and the European General Data Protection Regulation (GDPR) on the side of political regulation successfully went together toward citizen's rights, a similar strategy for ML models and regulations in concert is missing.

### 2.1. Inherent trustworthiness

From its beginning, the machine learning community aimed at offering users understandable machine learning processes and results. Interpretability guided the development of methods, their combination and transformation enabling users to inspect a learned model (Rüping, 2006). Recently, Chen et al. (2018) investigated interpretability of probabilistic models. Inductive logic programming (ILP) assumed that relational logic, and description logic in particular, is easily understandable (Morik and Kietz, 1991; Muggleton, 1991). Particular methods for interactive inspection and structuring of learned knowledge and given data offered a workbench for cooperative modeling of an expert with the ILP system (Morik, 1989). This close man-machine interaction in building a model creates common understanding of system developers and users. However, it does not scale to larger groups of affected stakeholders.

Decision trees were promoted as inherently understandable. However, feature selection and very deep trees quickly outgrow human intuition, and subtleties are only recognized by experts. For instance, the weights of redundant features are not appropriately computed in decision trees, whereas in support vector machines, they are (Mierswa and Wurst, 2005). Statistics, even if expressed in natural language, is not easy to understand correctly, as has been shown empirically (Wintle et al., 2019).

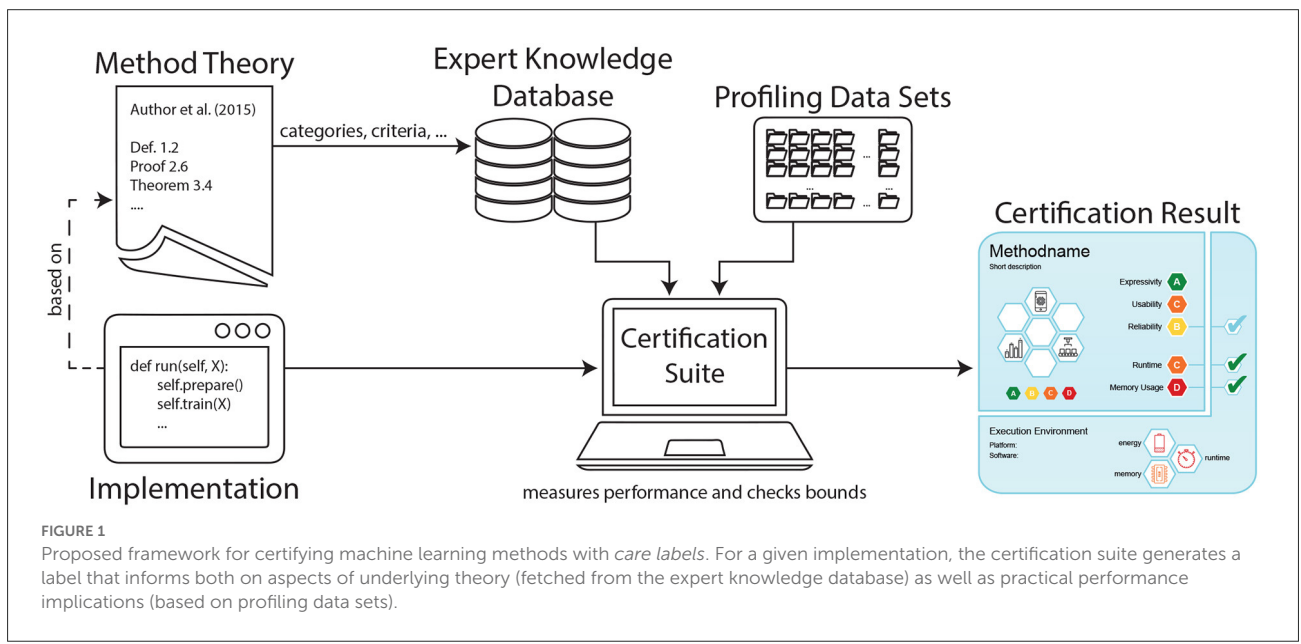

FIGURE 1
Proposed framework for certifying machine learning methods with *care labels*. For a given implementation, the certification suite generates a label that informs both on aspects of underlying theory (fetched from the expert knowledge database) as well as practical performance implications (based on profiling data sets).

## 2.2. Explainable AI

Explainable AI aims at offering an easy understanding of complex models for inspection by stakeholders. They investigate particular tasks, e.g., recommender systems (Nunes and Jannach, 2017), or particular ML families, e.g. Deep Neural Networks (Samek et al., 2019; Huang et al., 2020), or survey the needs of diverse stakeholders (Langer et al., 2021). Agnostic explanation routines are to explain a variety of learned models (Ribeiro et al., 2016; Guidotti et al., 2018). Given the large amount of research in this field, it has become a necessity in its own right to describe explanation methods along a proposed schema (Sokol and Flach, 2018). For model inspection by a domain expert or application developer, these methods are of significant importance.

## 2.3. Resource-aware machine learning

The approaches mentioned so far also do not consider resource usage (e.g., power consumption), even though it is crucial information for users (Henderson et al., 2020) and allows for discussing the environmental ethical impact of ML methods. Often, there is a trade-off between two important properties, such as higher runtime in exchange for lower energy consumption, which might influence the customer's decision on which model to choose. As an example, FPGAs offer unique performance and energy-saving advantages, but the software engineering part is challenging (Omondi and Rajapakse, 2006; Teubner et al., 2013). The challenges faced when deploying machine learning on diverse hardware in different application contexts (Hazelwood et al., 2018) even gave rise to a new conference, bridging the gap between machine learning and systems researchers (Ratner et al., 2019). The current increase in awareness regarding $CO_2$ emissions foregrounds these properties even more for users who want to design ML systems responsibly. Indeed, the amount of $CO_2$ emitted during training and deployment of state-of-the-art machine learning models has increased drastically in recent years (Schwartz et al., 2019; Strubell et al., 2020). To give more insight into this issue, Schwartz et al. (2019) urge researchers to provide a price tag for model training alongside the results, intending to increase visibility and making machine learning efficiency a more prominent point of evaluation. In Henderson et al. (2020), the authors provide a framework to measure energy consumption of ML models. However, they only measure specific model implementations, mostly disregarding theoretical properties and guarantees. We argue that a proper framework also needs to consider known theory, ideally stored as a database.

## 2.4. Description of methods and models

Modern machine learning toolboxes like RapidMiner, KNIME, or OpenML (Mierswa et al., 2006; van Rijn et al., 2013) are oriented toward knowledgeable users of ML or application developers. They and others (Brazdil et al., 2003; Falkner et al., 2018) use meta-data which offer a descriptive taxonomy of machine learning. In this sense, the ML processes are carefully described and documented. The user may click on any operator to receive its description and requirements. RapidMiner, in addition, recognizes problems of ML pipelines and recommends or automatically performs fixes. It enhances understandability by structuring ML pipelines and offering processes in terms of application classes.

Moving beyond the direct interaction between system and application developer aims at accountable descriptions of models and data. The approaches for *FactSheets* from IBM (Arnold et al., 2019) and *Model cards* from Google (Mitchell et al., 2019) are closely related to our approach. They give impetus to document particular models for specific use cases in the form of natural language and tabular descriptions, and even suggest to include them with ML publications.

In a recent user study, most interviewees found the idea of model-specific FactSheets helpful in understanding the methodology behind the model creation process (Hind et al., 2020). Another line of work aims at automatically tracking and visualizing the training process, including computed metrics as well as model architecture (Vartak et al., 2016; Schelter et al., 2017).

While these approaches are an important and necessary call for participation in the endeavor of describing learned models, we argue that natural language descriptions and empirical results alone are not enough to enhance trust in the model. They do not account for whether or not the theoretical properties of the model are fulfilled in the specific implementation at hand. This was stunningly shown by Dacrema et al. (2019), where baseline heuristics were able to beat top-tier methods, and many results could not be reproduced at all.

A summary of related work is shown in Table 1, highlighting which aspects of ML certification the authors cover, and where they fall short. What we find missing in the current state-of-the-art is a unifying concept to certify ML methods and their implementation on diverse hardware in terms of adherence to known theoretical properties and resulting resource demands. We argue that a method's properties need to be classified independently from a specific use case or data set. Additionally, this information needs to be accessible for non-experts, thus complicated theory has to be concealed through appropriate levels of abstraction. This is where our proposed *care labels* come into play, offering a comprehensive and easy to interpret

TABLE 1 Our proposed care label approach in comparison to other methods trying to certify certain aspects of the machine learning process.

|  | Arnold et al. (2019) | Mitchell et al. (2019) | Schwartz et al. (2019) | Henderson et al. (2020) | Care labels (this work) |
| --- | --- | --- | --- | --- | --- |
| Certify trained model | ✓ | ✓ | ✗ | ✓ | ✓ |
| Test compliance with theory | ✗ | ✗ | ✗ | ✗ | ✓ |
| Consider different hardware | ✗ | ✗ | (✓) | ✓ | ✓ |
| Comprehensible for non-experts | ✓ | ✓ | ✓ | ✗ | ✓ |
| Additional remarks | – | – | No practical application | Only focus on energy | – |

To the best of our knowledge, our approach is unique in combining practical measurements (like energy consumption) with checks of theoretical properties on different computing architectures. Brackets denote proposed methods that are not put into practice.

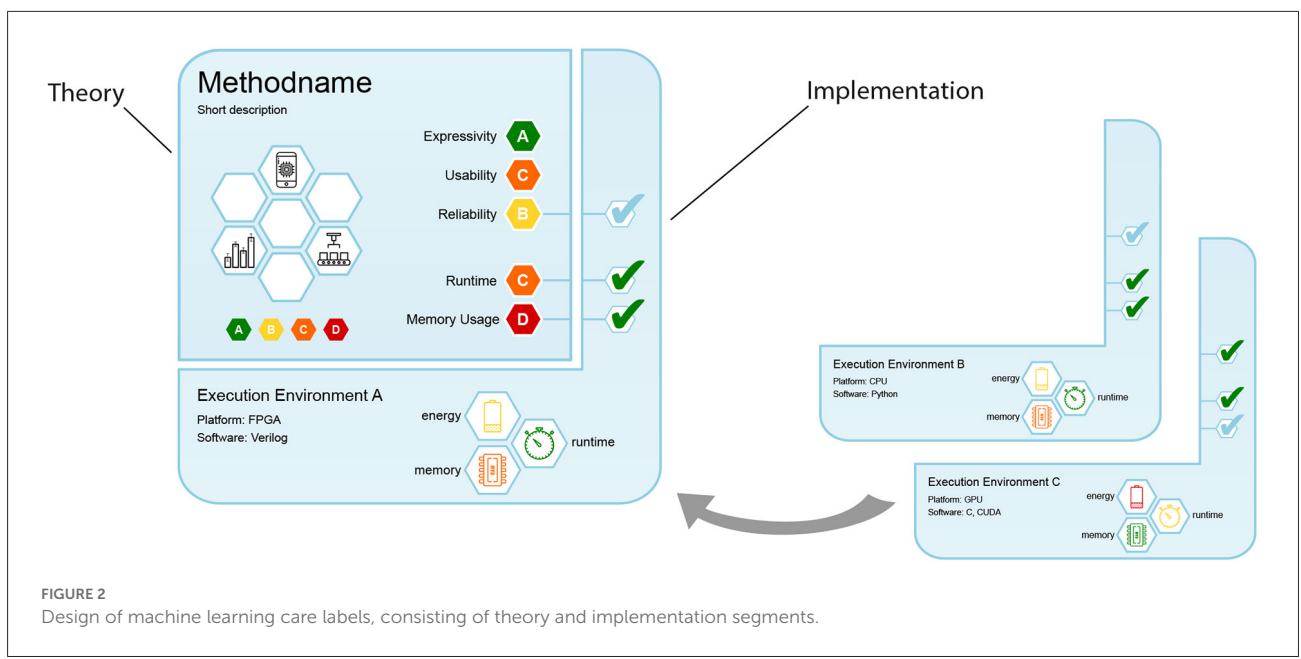

FIGURE 2
Design of machine learning care labels, consisting of theory and implementation segments.

overview of methodological properties, both theoretical and given the particular implementation at hand.

## 3. Machine learning care labels

We now introduce our care label concept for certifying ML methods, addressing all aforementioned issues. With "method," we refer to a combination of "components" for performing a specific machine learning task, such as training or applying a model. Most components are customizable, resulting in wildly varying properties. We later show this in practice for different probabilistic inference algorithms. Implementation and choice of hardware add yet another layer of complexity. Our care labels produce relief by hiding all underlying complexity behind a user-friendly façade. In the following sections, we discuss the details of our concept, following the structure of our contribution list from Section 1.

### 3.1. Care label design

Evidently, our proposed care labels need to take manifold theoretical and practical insights about ML methods into account, and compile them into a single short comprehensive document, similar to an index card. Our design is shown in Figure 2 and consists of two segments: The upper-left segment contains information about the method's theoretical properties, while the bottom-right segment also considers the given execution environment. As methods can be implemented in various ways and on various compute architectures, we designed this segment to "attach" to its theory. This is analogous to different brands of refrigerators: While the abstract task stays the same (keeping food and beverages cool), manufacturers use different components and circuits, and their specifics (e.g., lowest possible temperature, noise level, energy consumption) vary. In the same way, different ML implementations perform the same abstract task, but have their specific strengths and weaknesses.

Both segments contain a name and short description in their upper left corner. The theoretical segment displays the method's rating for five important *categories* (cf. Section 3.3) on the right, represented as colored hexagons. By restricting the care label to simple color-based ratings introduced in Section 3.2, we allow for a high-level assessment without the need for deeper understanding of the underlying theory. On the left, white hexagonal fields provide space for badges, which we describe in Section 3.5.

The segment contains three checkbox fields on the right that connect to three theoretical categories, indicating whether the theoretical properties are verified for the implementation. For a refrigerator, this could be a test whether the temperature reliably stays below the point where bacteria grow quickly. For ML methods, we can check if theoretical bounds about result quality, runtime or memory consumption hold.

Additionally, three white hexagonal fields with colored symbols at the bottom show measurement results for runtime, memory, and energy consumption (for more details see Section 3.4). In short, our design accomplishes the following:

- Provides general information about the ML method at a glance.
- Shows simple ratings for important categories, trading complexity and detail for simplicity and user-friendliness.
- Clearly highlights the interplay of theory and implementation by showing whether the implementation fulfills all theoretical properties.
- Is understandable for users without scientific background, allowing for easy comparison between ML methods.
- Highlights noteworthy properties that stakeholders may need for their particular application.

### 3.2. Expert knowledge database for method ratings

The theory behind machine learning methods is manifold, with different model classes having their own intricacies that can only be fully understood by experts. Consequently, they are required to assess important properties, identify to what extent a method exhibits them, and convey that knowledge to less informed users.

We propose to assemble a database with criteria that describe theoretical properties of ML methods, independent of their implementation. By bundling criteria into few concise categories, we allow for easier comparability between methods. We further propose to assign a rating to each category, consisting of four levels $A$—$D$, each represented by a color from a gradient ranging from green ($A$, best rating) to red ($D$, worst rating). This is inspired by similar certification concepts, e.g., the EU energy labeling framework (Council of European Union, 2017), which rates energy efficiency of electronic devices, or Nutri-Score for nutrient content in food (Julia et al., 2018). The ratings represent an expert assessment of how strongly the method at hand fulfills the respective criteria, as listed in Table 2.

As mentioned before, customizable components of a method can lead to quite different properties: While a basic model class may fulfill most criteria and consequently be rated $A$, a specific component may override the rating to $B$, e.g., due to much slower asymptotic runtime, or simplifying assumptions resulting in weaker theoretical guarantees. To address this issue, we propose a "building block" approach, by storing separate ratings for methods and their components in the database and combining them for a user-specified method. Where ratings for certain categories are not affected by configuring the component, we allow neutral ratings. Generally, ratings should be combined pessimistically, i.e. good ratings get overridden by worse ratings, this is depicted in Figure 3. This complexity is hidden from users, as they only receive a single label.

TABLE 2 Rating scale assigned to categories, depending on expert assessment, aided by proportion of fulfilled criteria.

| Letter | Symbol | Name |
|---|---|---|
| A |  | Method strongly and reliably exhibits the category's properties; all or most applicable criteria are met. |
| B |  | Method has many positive properties falling into this category; the majority of applicable criteria are met. |
| C |  | Method has few positive properties falling into this category, or has many positive properties but significant drawbacks; less than half of applicable criteria are met. |
| D |  | Method does not match the category's requirements at all, or significant drawbacks outweigh a few positive properties; no or very few applicable criteria are met. |

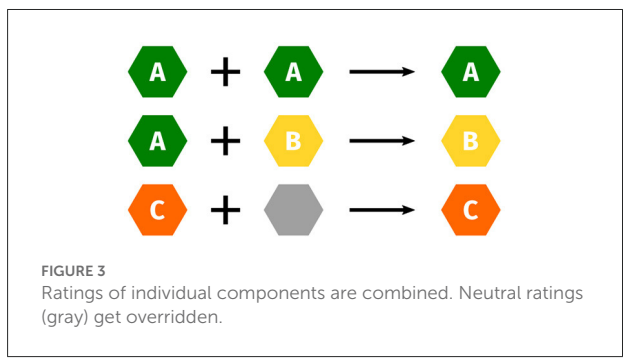

FIGURE 3
Ratings of individual components are combined. Neutral ratings (gray) get overridden.

## 3.3. Categories and criteria

In an attempt to untangle all aspects that ML users should consider when choosing a method and implementation, we propose a set of categories that summarize desirable properties.

For each category, we compile a list of yes-no type criteria: A high number of fulfilled criteria supports a method's aptitude w.r.t. the category, resulting in a higher rating. In this section we give only a few examples of criteria for each category, hoping that through practical insights and input from fellow researchers the list will grow.

While the categories we propose are designed to be more or less universal across model families, the constituting criteria may not apply to certain types of models or algorithms, e.g., because they belong to completely different paradigms. A solution to this problem would be to differentiate between model families, such as *generative* and *discriminative* models, providing alternative criteria for each separately.

We want to point out that we purposefully do *not* include categories or criteria concerning quality metrics like accuracy or convergence speed. These are highly dependent on specific input data and give no reliable impression of how well a method performs in an arbitrary, unknown use case. Giving simple ratings for these criteria would strongly contradict the "no free lunch" theorem—the fact that no single method performs universally well on all problems (Wolpert and Macready, 1997). We do however propose to investigate selected performance properties of method implementations, as described in Section 3.4.

### 3.3.1. Expressivity

- Example criterion: *The method provides at least one human-interpretable measure of uncertainty along with its output.*

We call a model *expressive* if it produces a variety of useful outputs for different input formats. This simple definition implies multiple properties: On the one hand, an expressive model should be able to handle arbitrarily complex functions, e.g., a classifier splitting every labeled data set, or a generative model approximating any probability distribution. On the other hand, highly expressive models provide additional outputs (e.g. measures of uncertainty or justification) to make them more interpretable. For users with certain safety-critical application contexts, such information may even be a strict requirement.

### 3.3.2. Usability

- Example criterion: *The model is free of hyperparameters.*

This category is concerned with the method's ease of use for end users. A main aspect is the number and complexity of hyperparameters: A hyperparameter-free method can be directly applied to a problem, while a method with many parameters needs fine-tuning to produce good results, which requires considerable effort and experience. Even more experience is required for choosing parameter values that are difficult to interpret: Choosing $k$ for a $k$-Means Clustering is conceptually much easier than choosing the weight decay in stochastic gradient descent. The difficulty of choosing optimal hyperparameters can be alleviated by theoretical knowledge of optimal parameter settings. We consider a method to be more easily usable if there are algorithms or formulas for deriving good or even optimal parameter values for given inputs.

### 3.3.3. Reliability

- Example criterion: *The method produces theoretically exact results.*

We require *reliable* models to be firmly grounded in theory, i.e. when there is evidence of mathematical error bounds, and if there are insights about the model's fairness or bias. As an example, uncertainty given by neuron activations in ANNs alone was found to not necessarily be a reliable measure (Guo et al., 2017). Such models are highly untrustworthy when they are used in safety-critical fields such as autonomous driving. Contrary, MRFs were proven to recover the correct marginal probabilities with increasing number of data points, given the underlying independence structure (Piatkowski, 2019). It is important to comprehensible visualize these fundamental differences.

Importantly, if there are theoretical bounds for the method at hand, they should also be verifiable by software tests, which we call *bound checks*. The particular tests need to be defined separately for all methods eligible for certification—we discuss details in Section 3.4.

### 3.3.4. Theoretical time and memory consumption

- Example criterion: *The method's runtime scales (at worst) quadratically with the input dimensionality, i.e. in $\mathcal{O}(n^2)$.*

Runtime and memory usage are factors of utmost importance for stakeholders, especially when facing resource constraints. ML theory provides insights on (worst-case) time and memory consumption of algorithms in the form of big $\mathcal{O}$ notation. Based on this theoretical tool, we propose a ranking of asymptotic time and memory complexity classes, with the rank being displayed in the care label's theory segment. In cases where big $\mathcal{O}$ notation depends on different factors (e.g., number of features or data points), we propose to classify the method according to the factor with the highest complexity class.

Energy is another important factor to consider when deploying ML, but it results directly from runtime, memory consumption and hardware. As such, theory does not provide any additional information here.

## 3.4. Certifying the implementation

So far we only considered static theoretical properties of ML methods, and how corresponding information can be summarized *via* simple care label ratings. However, looking at theory alone is not enough, as practically rolled out ML can diverge from it. Consider runtime as a practical example: In theory, an algorithm may be very efficient, but its implementation may still be very slow, due to slow periphery or inefficient code. Many popular ML implementations even suffer from severe bugs (Thung et al., 2012; Islam et al., 2019), let alone aligning with respective theoretical properties. We therefore propose to also certify the implementation's compliance with its underlying theory *via* test procedures that we call *bound checks*, which either investigate the dynamic aspects of *reliability* or *performance*.

The former intend to verify method-specific theoretical guarantees for reliability (cf. Section 3.3), as provided by the expert knowledge database. We propose to check those guarantees programmatically *via* software tests. This requires synthetic data with known properties, which is fed into the implementation. Its output is then checked against the known expected results.

Our performance checks investigate runtime and memory usage in the given execution environment. We here draw from the previously introduced asymptotic complexity classes, that the implementation is expected to comply with. We check this compliance by running experiments on synthetic data with varying input sizes. Measuring the corresponding runtime and memory usage in a software profiling fashion (cf. Section 4.2.2) allows for checking whether the theoretical complexity holds. Checkmark symbols on the right hand segment of our care label denote whether the implementation satisfies the reliability and performance checks.

Information about the available hardware allows to also assess the energy consumption, and thus, carbon footprint. As motivated earlier, this is of high interest for aspects of environmental ethics (Henderson et al., 2020). By drawing inspiration from electronic household devices and informing about energy consumption, our care label rewards energy-efficient implementations and ultra-low-power devices such as FPGAs and ASICs, even if they are limited in expressivity or have a higher runtime. For providing specific information, we assess the practical runtime (in seconds) and memory demand (in megabytes), along with energy consumption (in Watt-seconds) for a medium-sized data set. Those measurements are displayed in the implementation segment. For comprehensibility we also display them as colored badges, based on their position on a scale for different orders of magnitudes.

## 3.5. Badges for noteworthy criteria

We argue that there are certain properties, which are particularly noteworthy, because they are rare among comparable methods and have great impact on the method's overall rating. Examples of such noteworthy properties include uncertainty measures, whether the method can be tested for robustness, and whether the model can be used with streaming data. In order to highlight these properties, we introduce badges in form of pictograms that get printed on the care labels. Some examples of badges, along with short explanations, are given in Figure 4.

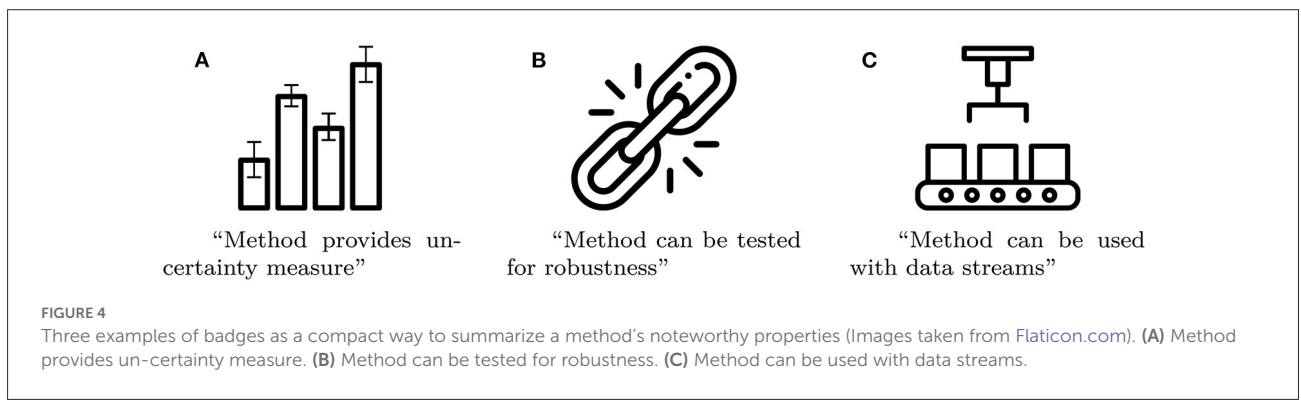

FIGURE 4
Three examples of badges as a compact way to summarize a method's noteworthy properties (Images taken from Flaticon.com). **(A)** Method provides un-certainty measure. **(B)** Method can be tested for robustness. **(C)** Method can be used with data streams.

## 3.6. Certification suite concept

We propose to develop a certification suite software that enables a less informed user to enquire comprehensible information on ML methods and their implementations, in the form of care labels. Most importantly, it allows to configure a specific method from its available constituting components. For the chosen configuration, the software queries the ranking information, asymptotic performance bounds, and reliability bounds from the knowledge database, and combines them into the theoretical label segment. After configuring the method's backend implementation according to the user input, the suite then profiles and runs bound checks. We also propose to implement an interactive, high-level interface, that hides all complicated ML logic from users who are not very experienced in the field of ML. In terms of our work we have already implemented a simple prototype, see Section 4.2.2 for more information.

## 4. Applying the care label concept to graphical models

We now implement our concept for selected members of the probabilistic graphical model (PGM) family. Having a long history in statistics and computer science, theoretical properties of graphical models have been studied rigorously in literature (Koller and Friedman, 2009). Thus, they are well-suited to demonstrate our care label concept as a means to aid the user in their decision making process. Firstly, we briefly discuss the theoretical background of Markov random fields (MRFs), a specific subtype of probabilistic graphical models (PGMs). Secondly, we present the care label generation procedure for two different MRF variants (Section 4.2). We discuss the static, theoretical properties and corresponding rating, as stored in the expert knowledge database, determining the theory segment of our label (Section 4.2.1). In addition, we present results of our testing procedures, which certify a given MRF implementation against the underlying theory, both in terms of reliability and resource demand, while also assessing the energy consumption of the execution environment at hand (Section 4.2.2).

## 4.1. Background on Markov random fields

MRFs belong to the family of PGMs and are used in many different applications like satellite image gap filling (Fischer et al., 2020), medical diagnosis (Schmidt, 2017), and security-critical systems (Lin et al., 2018). Moreover, MRFs can be used for constrained learning scenarios, like distributed environments (Heppe et al., 2020) or platforms under strict resource requirements (Piatkowski et al., 2016). We now shortly discuss the underlying theory of MRFs, as it provides guarantees and static properties that determine their care label ratings.

MRFs combine aspects from graph and probability theory in order to model complex probability distributions $\mathbb{P}(X)$ over some $d$-dimensional random vector $X = (X_1, \ldots, X_d)^\top$ efficiently. Conditional independences between elements of $X$ are exploited and modeled through a graph $G = (V, E)$, where each vertex $v \in G$ is associated with one random variable of $X$. If two vertices $i$ and $j$ are not connected by an edge, $X_i$ and $X_j$ are conditionally independent given the remaining variables. In this work we focus on discrete MRFs, allowing for an intuitive parametrization, therefore each element $X_i$ can take values in its discrete finite state space $\mathcal{X}_i$. By introducing the so-called potential functions $\psi_C : \mathcal{X}_C \mapsto \mathbb{R}_+$ for each of the cliques in $G$, mapping variable assignments to positive values, the joint density factorizes according to Hammersley and Clifford (1971) as follows:

$$\mathbb{P}(X = x) = \frac{1}{Z} \prod_{C \in \mathcal{C}(G)} \psi_C(x_C) \;, \qquad (1)$$

where Z, which is called *partition function*, acts as normalization,

$$Z = \sum_{x \in \mathcal{X}} \prod_{C \in \mathcal{C}(G)} \psi_C(x_C). \qquad (2)$$

The potential functions $\psi$ are parametrized by weights, which allows to define a *loss function*. By minimizing it during the training process the model adapts to a given data set. Typically, the weights are learned *via* maximum-likelihood-estimation with first order *optimization methods*.

Having access to the joint density clears the path for many further ML tasks, such as generating data *via* Gibbs sampling, answering marginal $\mathbb{P}(X_i = x)$ or conditional $\mathbb{P}(X_i = x_i | X_j = x_j)$ probability queries, or providing maximum a-posteriori (MAP) estimates. However, solving such tasks requires *probabilistic inference*. Algorithms for such computations can be divided into two variants, namely *exact* and *approximate* algorithms (see upper part of Figure 5). On the one hand, the junction tree (JT) algorithm (Wainwright and Jordan, 2008) is a well-known method to perform exact inference on arbitrary graph structures, while having a very slow asymptotic runtime. On the other hand, the loopy belief propagation (LBP) algorithm (Kim and Pearl, 1983) performs approximate inference, sacrificing theoretical guarantees for considerably faster performance. Further approximation algorithms include the variational (Wainwright et al., 2003) and sampling-based approaches (Andrieu et al., 2003). Keep in mind that exact probabilistic inference is a #P-complete task (Piatkowski, 2018).

## 4.2. Deriving care labels for Markov random fields

We generate care labels for different MRFs, based on combinations of chosen components. In the context of MRFs, there are three major configurable components: an *optimizer*, a *loss function*, and an *inference algorithm*. We restrict ourselves to investigating *gradient descent* optimization with a *likelihood* loss function, using either the LBP or the JT algorithm for performing inference. These combinations are depicted in Figure 5, along with their corresponding final care labels.

### 4.2.1. Expert knowledge-based ratings

MRFs already exhibit certain static properties and receive corresponding ratings for our categories. The ratings in the theoretical segment of the care label are stored in the expert knowledge database (cf. Section 3.2) and were agreed upon by 10 experts using a majority vote. In case of a tie, we decided in favor of the method. As combining the individual components can greatly influence the rating, we also have to identify their individual ratings. Combining the component ratings should be based on a fixed set of rules. Here, we stick to the already proposed way of taking the infimum overall ratings (cf. Figure 3). We display the respective ratings for the MRF components and variants in Table 3, and now explain their expert-knowledge-based justification and corresponding implications for the user, following the columns from left to right.

#### 4.2.1.1. Expressivity

Looking at Table 3, it stands out that the general ML method choice is the only component affecting the expressivity rating. We reason that the expressive power of MRFs is determined only by its inherent properties, while the customizable components are neutral. MRFs are very expressive, because they can perform all generative model tasks: they can be queried for conditional or joint probabilities and they allow to sample data from the distribution. In addition, *their probability output is a natural uncertainty measure*. Therefore, we argue that MRFs should be rated *A*.

#### 4.2.1.2. Usability

In terms of usability, MRFs receive the grade *B*, since the independence graph is usually unknown for real-world use cases, and thus has to be defined manually. For this, the user must either incorporate expert knowledge or use procedures for structure estimation (Yang et al., 2014). The loss function does not impact the usability and is rated neutral. Gradient descent only requires choosing a step size, which is well-documented with reasonable defaults, therefore we rate it *A*. The LBP inference requires careful tuning of the stopping criterion by specifying the convergence threshold or number of iterations. However, more iterations do not necessarily improve performance, which makes the choice quite unintuitive, resulting in a *C* rating for usability. JT inference does not require additional hyperparameter tuning, which yields an *A* rating. The usability rating shows the user that LBP makes MRFs a bit harder to use.

#### 4.2.1.3. Reliability

When provided with an exact inference algorithm and a convex optimization problem, MRFs are guaranteed to recover the correct distribution, with the error being bounded by the number of training instances (Bradley and Guestrin, 2012). This bound, which we call *distribution recovery check*, can be verified through software, resulting in an *A* rating. Since the likelihood as a loss function exhibits strong statistical guarantees, like *consistency*, *unbiasedness* and *efficiency*, we also award it with *A*. Given a density, which is convex w.r.t. parameters, a gradient descent optimizer is able to *recover the global optimum*, resulting in an *A* rating. This is also verifiable for the dynamic properties of a specific implementation *via* software, the so-called *convergence check*. However, all this reliability is only given if the chosen inference algorithm provides exact results for the gradient update. We reflect this restriction by assigning *D* and *A* ratings to LBP and JT respectively, as the former does not even provide a bound on the approximation error[2]. Our ratings clearly show the user that LBP makes MRFs a lot less reliable.

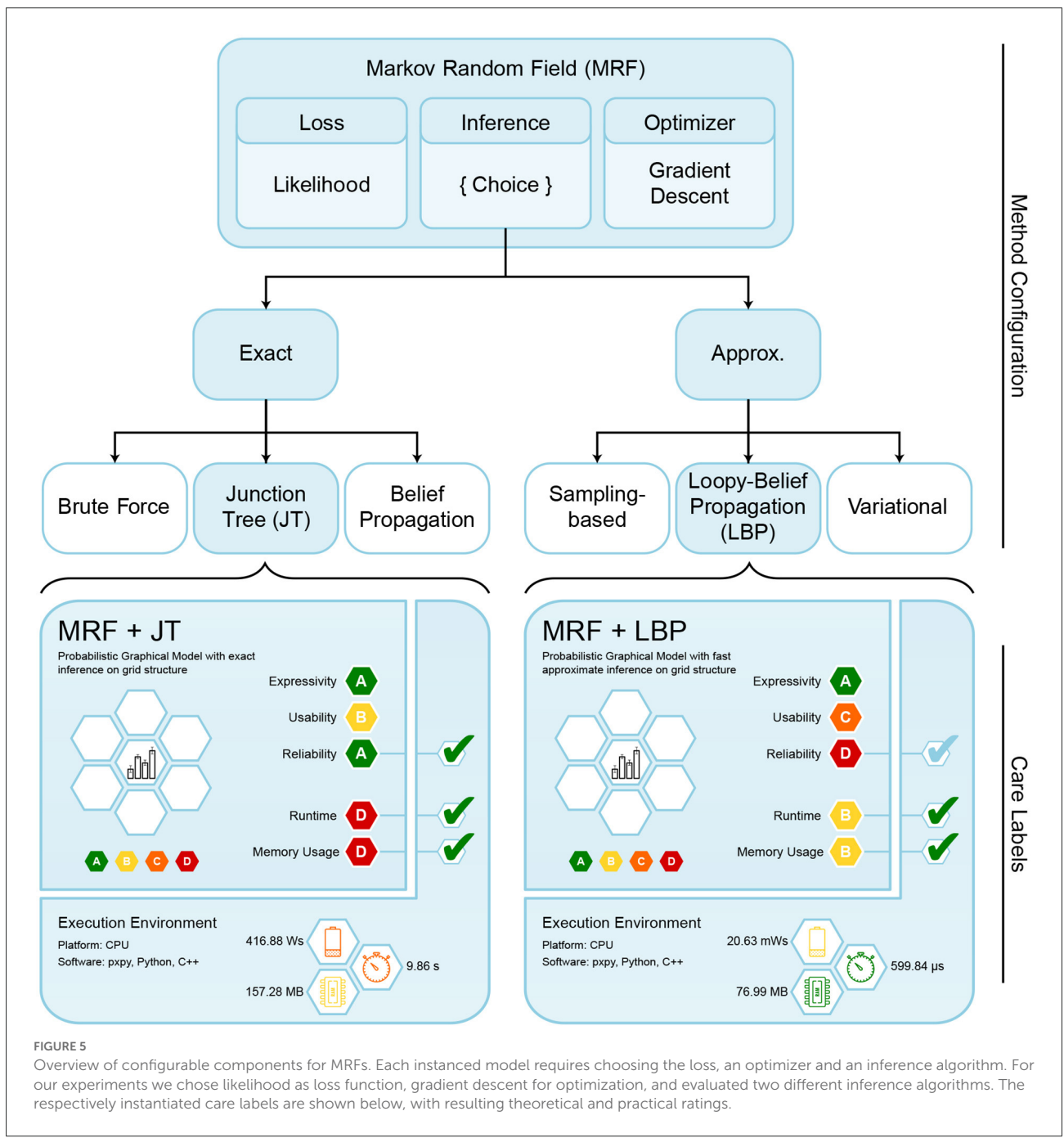

FIGURE 5
Overview of configurable components for MRFs. Each instanced model requires choosing the loss, an optimizer and an inference algorithm. For our experiments we chose likelihood as loss function, gradient descent for optimization, and evaluated two different inference algorithms. The respectively instantiated care labels are shown below, with resulting theoretical and practical ratings.

### 4.2.1.4. Runtime and memory

The theoretical runtime and memory complexity classes depend mostly on the chosen inference algorithm component,

---

2 Under certain, special conditions LBP may have guarantees, but this is generally not the case (Ihler et al., 2005).

while the MRFs type's general runtime is unspecified. Their memory demand for parametrization depends on the data complexity in two ways: It grows linearly with the number of features and quadratically with the number of discrete states, resulting in a $B$ rating. Due to only requiring a plain mathematical function evaluation, the likelihood loss acts neutrally on both categories. The gradient descent optimizer itself is resource efficient, assuming we are given the gradients of our loss function. Both memory and runtime complexity scale

TABLE 3 Ratings for individual components based on theoretical properties and expert knowledge rated from *A* to *D* alongside the final combined ratings for two concrete configurations.

| Component | Choice | Expressivity | Usability | Reliability | Runtime | Memory |
|---|---|---|---|---|---|---|
| ML Method | Markov random field | A | B | A | | B |
| Loss | Likelihood | | | A | | |
| Optimizer | Gradient descent | | A | A | A | A |
| Inference | Loopy belief propagation | | C | D | B | A |
|  | Junction tree | | A | A | D | D |
| Care labels | MRF + LBP | A | C | D | B | B |
|  | MRF + JT | A | B | A | D | D |

Categories for which the criteria do not apply are shown in gray and are treated neutral toward the final rating. The final ratings are typically derived from the worst-case of all rated components, but may depend on additional expert knowledge instead.

TABLE 4 Guarantees and resource consumption of different inference algorithms.

| | Junction tree | LBP |
|---|---|---|
| Time complexity | $O(X_{C_{max}}^w)$ | $O(I|E|X_{max}^2 N_{max})$ |
| Memory complexity | $\sum_{C \in C(JT)} \prod_{v \in C} X_v$ | $2. \sum_{(s,t) \in E} |X_s| + |X_t|$ |
| Exact | Yes (Lauritzen and Spiegelhalter, 1988) | Heuristic (Murphy et al., 1999) |

linearly with the model dimension, resulting in an *A* rating for both categories.

The resource requirements of our inference algorithms are shown in Table 4 (Piatkowski, 2018). We rated the runtime of JT inference with *D* due to scaling exponentially with the junction tree's width[3] *w*. Memory receives the same rating, because the underlying data structures also grow exponentially with tree width. Thus, users might find JT inference feasible for sparse independence structures, while for dense graphs its runtime and memory demand may exceed available resources. LBP on the other hand works efficiently on general graphs and even provides exact solutions for trees and polytrees. It is rated *B* due to scaling quadratically in number of states of the largest state space $\mathcal{X}_{max}$, and linearly in the number of edges $|E|$ in the graph, the number of iterations $I$ and the size of the largest neighborhood $\mathcal{N}_{max}$ in $G$. The memory consumption for LBP is rated *A* as we only have to store intermediate results, whose memory demands scale linearly in the number of states per clique. Choosing between JT and LBP inference, the user can trade exactness and strong guarantees for better runtime and memory. This is illustrated in Figure 7, showing runtime and memory consumption for MRFs with increasing number of vertices.

### 4.2.2. Testing the implementation

To test the dynamic properties of specific MRF implementations and derive the care label's implementation segments, we implemented a certification suite, as described in Section 3.6. It draws the theory-based static bounds and ratings from the expert knowledge database, performs reliability bound checks, investigates the implementation's behavior in terms of runtime, memory and energy consumption, and outputs the complete care label.

#### 4.2.2.1. Experimental setup

In order to run our bound checks (cf. Section 3.4) we generated synthetic data sets by defining specific distributions and sampling from them. Having access to both the sampled data and their underlying distribution parameters allows for assessing whether reliability checks pass. As graph structure we chose a grid graph, with binary state space and increasing grid size from $2 \times 2$ up to $15 \times 15$. This resulted in data sets of different sizes, which have been utilized to perform the resource bound checks.

---

3 The junction tree (Huang and Darwiche, 1996) is an auxiliary graph which needs to be derived in order to run the complete inference algorithm.

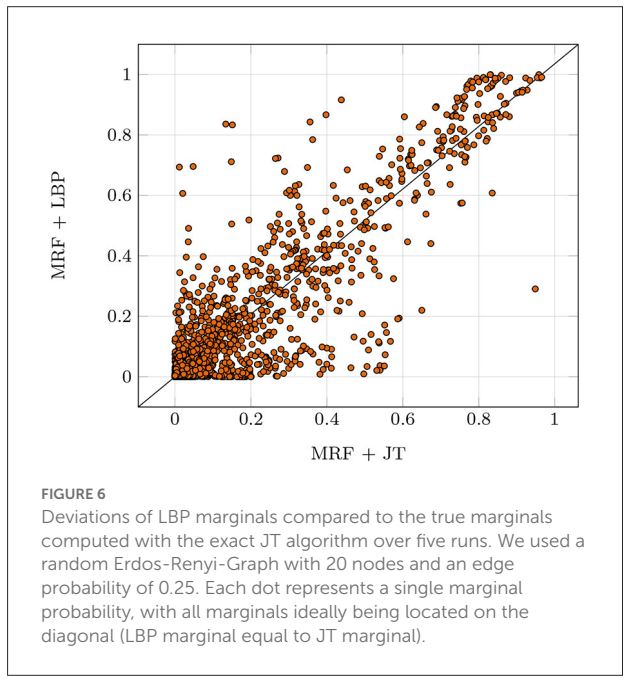

FIGURE 6
Deviations of LBP marginals compared to the true marginals computed with the exact JT algorithm over five runs. We used a random Erdos-Renyi-Graph with 20 nodes and an edge probability of 0.25. Each dot represents a single marginal probability, with all marginals ideally being located on the diagonal (LBP marginal equal to JT marginal).

For running the MRFs logic, we utilized the pxpy[4] library, which implements JT and LBP.

Our certification suite measures runtime, memory demand[5], and CPU energy consumption[6] *via* established tools, similar to Henderson et al. (2020). All experiments were performed on a workstation equipped with an Intel(R) Xeon(R) W-2155 CPU, 64 GB RAM and Ubuntu 20.04.1 LTS as operating system.

#### 4.2.2.2. Reliability bound checks

To verify the reliability of the implementation against the static characteristics described in Section 4.2.1, which are especially important for safety-critical applications, we perform two exemplary bound checks: The *distribution recovery check* (Hoeffding, 1963) and the *likelihood convergence check*. The first check is performed by comparing the true marginal probabilities $\mu^*$ to the marginals $\hat{\mu}$ computed by the provided implementation for the true parameters $\theta^*$. Therefore, cliquewise KL-divergences are computed and reduced to the max value. If this value falls below a given threshold, the check passes. The second check verifies if the given implementation fits the data. To this end, we run an optimization procedure with the true structure and our samples, checking convergence based on the gradient norm. If the norm falls below a given threshold, the check passes. The investigated JT implementation was able to pass both checks. Recall that LBP inference is not exact as depicted in Figure 6 and received a *D* for reliability. Even though the LBP algorithm does not exhibit theoretical guarantees, it was able to pass the reliability tests for some data sets. Still, it failed for most, therefore the implementation did not receive the reliability checkmark.

---

[4] https://pypi.org/project/pxpy/
[5] https://github.com/giampaolo/psutil
[6] https://github.com/wkatsak/py-rapl

#### 4.2.2.3. Runtime and memory bound checks

Next, we evaluate whether the performance in the given execution environment complies with the identified complexity classes. We depict the measured resource consumption for both JT and LBP configurations in Figure 7. As expected, it shows that both memory usage and runtime increase with the number of vertices for JT. For our automatic checks, we fit different linear regression models with the resource measurements (i.e. one model for linear, quadratic, cubic, etc. complexity). We also cross-validated this assessment by subdividing the measurements into several independent sets, and fitting the regression for each group. In our experiments, those results corresponded to the identified theoretical complexity of the tested MRF configurations, thus all methods receive memory and runtime checkmarks.

#### 4.2.2.4. Resource consumption testing

For specific resource measurements, we chose a medium-sized data set stemming from a grid-structured graph with $14 \times 14$ vertices and binary state space. The results are displayed in Table 5. They confirm that the JT configuration requires much more runtime and energy than LBP. The hardware platform was internally measured to consume an average of $20 - 43$ Watt per experiment. To obtain the complete energy consumption, we multiplied the power with runtime. The badge colors in the implementation part of the care labels are directly derived from those measurements.

Our experimental findings show the usefulness of our care label concept, compacting the extensive theory of PGMs, while still providing useful information that is otherwise not accessible for users.

## 5. Conclusion and further work

With state-of-the-art systems, ML user requirements can differ vastly. Certain users might know the theory or have an intuitive understanding of properties and guarantees. Often, however, users are not aware of the intricacies of different methods. There are approaches that discuss how trust in ML can be increased, but they often fail to connect theory and practice, or are too abstract and inaccessible to non-experts who do not want to understand system in the same manner that they do not want to understand their washing machine.

We address these issues *via* our care labels to inform a broad range of users and ML customers. Our labels identify theoretical properties that are highly relevant for safety-critical or resource-constrained use cases. We test implementations against theory by performing bound checks for reliability and

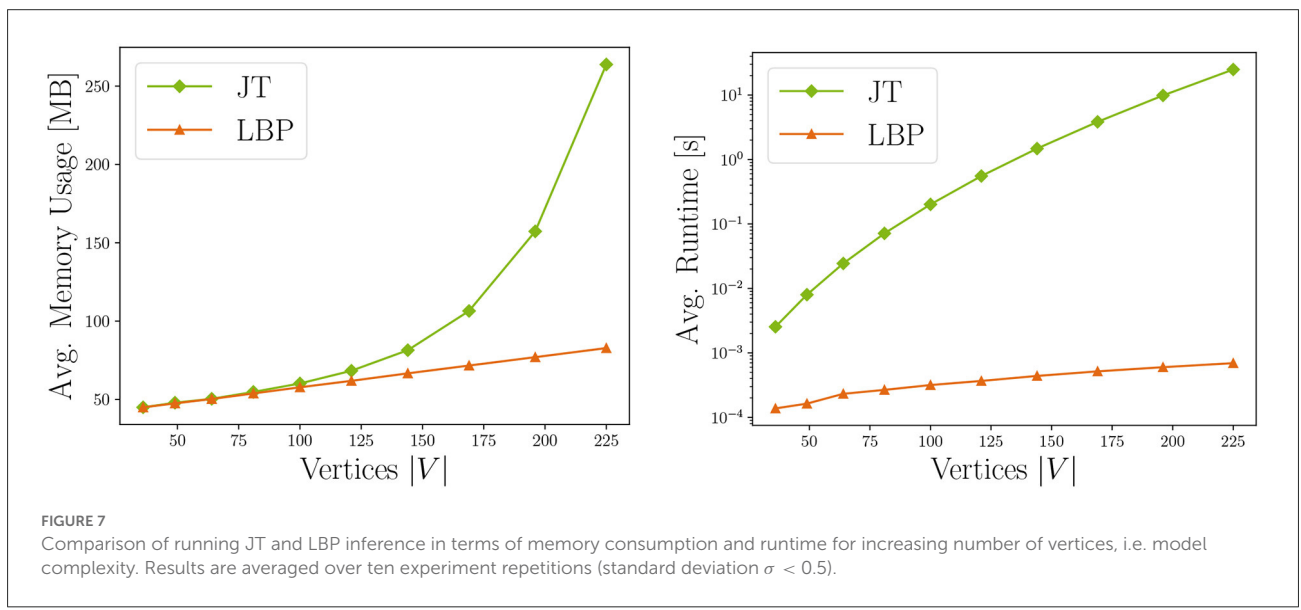

FIGURE 7
Comparison of running JT and LBP inference in terms of memory consumption and runtime for increasing number of vertices, i.e. model complexity. Results are averaged over ten experiment repetitions (standard deviation $\sigma < 0.5$).

TABLE 5 Comparison of the two methods on a data set with a grid independence structure sized 14 × 14.

| Measurements | JT | LBP |
| --- | --- | --- |
| Runtime | 9.86 s | 599.84 µs |
| Memory | 157.28 MB | 76.99 MB |
| Energy | 416.88 W s | 20.63 mW s |

The experiments were repeated ten times with a standard deviation of $\sigma < 0.5$.

measuring resource consumption. All this information is neatly displayed in our care label design, which is easy-to-understand for both experts and customers without a scientific background. If demanded, more intricate details could be provided to users.

We demonstrated that our concept is practical for MRFs as an example for undirected generative models. For the experimental evaluation, we implemented a verification suite, expert knowledge database, and care label design (cf. Figure 1). Looking at their inference, we have inspected the exact JT and the approximate LBP algorithms. The generated labels allow users to assess implications of using MRF variants with different components where the extensive amount of theory behind PGMs remains invisible to the user.

Subsequent to this work, we intend to refine and finally publish a verification suite for ML practitioners. Here, we contributed the general framework concept and proof-of-concept results for MRFs. Future work could investigate other probabilistic inference methods, i.e. variational inference and the MCMC method. Tests for discriminative methods like Conditional Random Fields or directed PGMss like Latent Dirichlet Allocation or Hidden Markov Models are yet to be generated in order to assign care labels to these.

Other ML methods like deep neural networks might also bring attention to totally different properties like robustness. A wide range of tools for testing robustness or measuring resource consumption is already available, e.g., cleverhans https://github.com/cleverhans-lab/, carbon tracker https://github.com/lfwa/carbontracker and many more. In order to integrate those tools into our framework, the tests become related to a particular architecture with a particular parameter setting. This is more complex, but preliminary studies already show that it is possible.

The question of scalability is indeed a pressing one, because the expert knowledge database and criteria checks need to be assembled and implemented manually. The ultimate goal would be to automatically generate tests from the proofs and experiments that are published in scientific papers. The collection of https://paperswithcode.com is a first step into that direction. Turning this into generating calls of executing experiments whose results can be framed as care labels is the very long-term perspective of our approach.

## Data availability statement

The original contributions presented in the study are included in the article/supplementary material, further inquiries can be directed to the corresponding author.

## Author contributions

KM had the idea with the care labes and KM and HK designed the study. DH, RF, and AP performed the statistical analysis. LH, DH, RF, SM, AP, and MJ wrote the first draft of the manuscript. KM, HK, and NP wrote sections of the manuscript. All authors contributed to manuscript revision, read, and approved the submitted version.

# Funding

This research has been funded by the Federal Ministry of Education and Research of Germany and the state of North-Rhine Westphalia as part of the Lamarr-Institute for Machine Learning and Artificial Intelligence, LAMARR22B. Parts of the work on this paper has been supported by Deutsche Forschungsgemeinschaft (DFG)—project number 124020371—within the Collaborative Research Center SFB 876 'Providing Information by Resource-Constrained Analysis', DFG project number 124020371, SFB project A1.

# Conflict of interest

The authors declare that the research was conducted in the absence of any commercial or financial relationships that could be construed as a potential conflict of interest.

# Publisher's note

All claims expressed in this article are solely those of the authors and do not necessarily represent those of their affiliated organizations, or those of the publisher, the editors and the reviewers. Any product that may be evaluated in this article, or claim that may be made by its manufacturer, is not guaranteed or endorsed by the publisher.